\documentclass[journal]{IEEEtran}
\usepackage{cite}
\usepackage{amsmath,amssymb,amsfonts}
\usepackage{algorithmic}
\usepackage{graphicx}
\usepackage{textcomp}
\usepackage{booktabs}
\usepackage{multirow}
\usepackage{times}
\usepackage{epsfig}
\usepackage{soul}
\usepackage{mathtools}
\usepackage{dsfont}
\usepackage{tabularx}
\usepackage[table,xcdraw]{xcolor}
\usepackage{enumitem}

\definecolor{selectiveyellow}{rgb}{0.965, 0.757, 0.259}
\definecolor{dodgerblue}{rgb}{0.412, 0.607, 0.816}
\definecolor{lightgreen}{rgb}{0.494, 0.666, 0.333}

\begin{document}

\title{PC-SwinMorph: \\ Patch Representation for Unsupervised Medical Image Registration and Segmentation}

\author{Lihao Liu, Zhening Huang, Pietro Liò, Carola-Bibiane Schönlieb and Angelica I Aviles-Rivero
\thanks{L. Liu., C. Schönlieb, and A.I. Aviles-Rivero are with the Department of Applied Mathematics and Theoretical Physics, University of Cambridge. Cambridge CB3 0WA, UK. Corresponding author: ll610@cam.ac.uk.}
\thanks{Z. Huang. is with the Department of Engineering, University of Cambridge. Cambridge CB2 1PZ, UK.}
\thanks{P. Liò. is with the Department of Computer Science, University of Cambridge. Cambridge CB3 0FD, UK.}
}

\maketitle

\begin{abstract}
Medical image registration and segmentation are critical tasks for several clinical procedures. Manual realisation of those tasks is time-consuming and the quality is highly dependent on the level of expertise of the physician. To mitigate that laborious task, automatic tools have been developed where the majority of solutions are supervised techniques. However, in the medical domain, the strong assumption of having a well-representative ground truth is far from being realistic. To overcome this challenge, unsupervised techniques have been investigated. However, they are still limited in performance and they fail to produce plausible results. In this work, we propose a novel unified unsupervised framework for image registration and segmentation that we called PC-SwinMorph. The core of our framework is two patch-based strategies, where we demonstrate that patch representation is key for performance gain. 
We first introduce a patch-based contrastive strategy that enforces locality conditions and richer feature representation. We also introduce a novel patch stitching strategy based on a 3D window/shifted-window multi-head self-attention module to eliminate artifacts from the patch splitting. We demonstrate, through a set of numerical and visual results, that our technique outperforms current state-of-the-art unsupervised techniques.
%We demonstrate enforces more meaningful feature representation whilst enforcing local and global structure representation.

\end{abstract}

\begin{IEEEkeywords}
Patch Representation, Unsupervised Learning, Image Registration, Image Segmentation.
\end{IEEEkeywords}

%%%%%%%%% BODY TEXT
\section{Introduction} \label{sec:intro}
Image registration is a fundamental task in medical image analysis, which aims at finding a mapping that aligns an unaligned image to a reference one. {The estimated spatial mapping (deformation field)} seeks to best align the anatomical structure of interest. Image registration is relevant for several tasks in clinical practice including image-guided surgery~\cite{aviles2016towards, han2021fracture}, segmentation~\cite{fu2017automatic,liu2019probabilistic} and image reconstruction \cite{liu2021rethinking}. The outcome of those tasks greatly depends on the quality and efficiency of the registration technique. Although traditional image registration techniques~\cite{rueckert1999nonrigid, vercauteren2009diffeomorphic, beg2005computing,hart2009optimal} are able to generate a good mapping between images, they build upon costly optimisation schemes, which limits their efficiency when using a large volume of data. With that limitation in mind, several deep learning techniques have been proposed for registration.

A major category of approaches is supervised image registration techniques~\cite{yang2017quicksilver,sokooti2017nonrigid,rohe2017svf,cao2017deformable,cao2018deformable}, where a good quality ground-truth is required for training. However and unlike other tasks in image analysis, it is very difficult to obtain high-quality {ground-truth deformation fields or segmentation masks}. Although a good mapping can be obtained from traditional methods or using synthetic data, this drawback hinders the performance and feasibility of those techniques in clinical practice. 

% \begin{figure}[t!]
%   \centering
%   \includegraphics[width=\linewidth]{fig/teaser1c.pdf}
%   %\includegraphics[width=0.5\textwidth]{fig/teaser1c.pdf}
%   %\includegraphics[width=6cm]{example-image-a}
%   \centering
%   \caption{Workflow overview of our PC-SwinMorph framework. We highlight our two patch-based strategies. For strategy I, we introduce a patchwise contrastive solution, which enforces a better feature representation. Whilst for strategy II, we highlight a multi-head self-attention module for stitching the patches.
%   }
%   \label{fig:introduction_network}
% \end{figure}

To mitigate the aforementioned strong requirement of supervised methods, a body of literature has been devoted to developing unsupervised techniques~\cite{mok2020fast,liu2020contrastive,kim2021cyclemorph,ye2021deeptag}. Those techniques have proposed different network mechanisms and explicit regularisers embedded in the architectures to enforce better correspondences between images. However, unsupervised techniques are still limited in performance compared to supervised methods. This is due to the lack of high-quality prior knowledge that introduces challenges such as failure in long-range correspondences. With the aim to alleviate this problem, recent techniques yet scarce have used vision transformers (ViT), where the self-attention mechanism~\cite{chen2021vit,zhang2021learning} is key for improving the correspondence of the image. Another strategy reported for unsupervised image registration is the use of contrastive mechanisms~\cite{liu2020contrastive} for improving feature representation, and therefore, enforcing a better mapping estimation between images. Although these techniques have reported improved performance for unsupervised image registration, it is still limited.% \lihao{Maybe not to mention supervised method anymore, because our method didn't outperform supervised methods in the end} and fine details are not preserved.

In this work, we proposed a unified framework for unsupervised image registration and segmentation, which we call PC-SwinMorph (\textbf{P}atch \textbf{C}ontrastive Strategy with \textbf{S}hifted-\textbf{win}dow multi-head self-attention). For a fair comparison to the state-of-the-art techniques, we use as backbone Voxel\textbf{Morph}. Medical images are more complex than natural images due to the anatomical structures, where fine details are of clinical relevance. \textit{We then hypothesise that patch embeddings are a more meaningful representation for performance gain in medical data. This is due to the spatial structure of the patch that allows capturing not only global but, more importantly, also local anatomical representations. Our PC-SwinMorph then enforces more meaningful feature representation whilst enforcing local and global structure representation. } Our contributions are as follows: %\vspace{-0.2cm}

%In this work, to solve the above challenges and come up with a more refined registration result, we propose to perform registration based on patches. 
%
%Medical images normally have more complex architectures, such as the curve and sulcus in brain images.
%
%Hence, we argue that we can get a better registration result by focusing on the alignment between two small patches.
%
%Specifically, we propose an \textbf{P}atch \textbf{C}ontrastive Framework with \textbf{S}hifted-\textbf{win}dow Multi-head Self-attention based on Voxel\textbf{Morph} for unsupervised 3D medical image registration and segmentation that we called Patch Contrastive-SwinMorph (PC-SwinMorph). The basic workflow of our framework is illustrated in Figure~\ref{fig:introduction_network}.
%Our contributions are as follows:

\begin{itemize}[noitemsep]
  \item We propose a patch-based framework for unsupervised image registration and segmentation, in which we highlight a patch-based contrastive strategy for enforcing a better fine detailed alignment and richer feature representation.
    %for unsupervised medical image registration and segmentation, in which we highlight the patch-level registration 
    
    %patch-level contrastive learning for a better
    
  \item We introduce a novel patch stitching strategy to alleviate the splitting effect caused by the patch representation. To do this, we use the 3D window/shifted-window multi-head self-attention module (3D W-MSA and 3D SW-MSA) to enable information exchange between different patches.
    %to refine the splitting effect from stitching patches.
  % \end{itemize}
  %\item We proposed a patch-based contrastive framework for unsupervised medical image registration and segmentation, in which we highlight the patch-level registration for detail alignment, and patch-level contrastive learning for better feature learning.
  %\item We proposed a patch stitching strategy to alleviate the splitting effect caused by direct stitching. We use the 3D Window / Shifted-window Multi-head Self-attention module to enable information exchange between different patches to refine the splitting effect from stitching patches.
  
  \item We evaluate our framework using two major medical benchmark datasets CANDI and LPBA40. We demonstrate from the numerical and visual results that our two patch-based strategies lead to better performance than the state-of-the-art techniques for unsupervised registration and segmentation.
  
  %for unsupervised registration and segmentation, we can largely improve the unsupervised registration and segmentation performance. 
  %our proposed methods with other 4 state-of-the-art methods on two major benchmark datasets CANDI and LPBA40. 
  
\end{itemize}

\section{Related Work}
\label{sec:related_work}
The problem of image registration has been extensively investigated in the literature, in which solutions broadly divide into classic techniques e.g.~\cite{rueckert1999nonrigid, vercauteren2009diffeomorphic, beg2005computing,hart2009optimal} and learning-based methods e.g.~\cite{yang2017quicksilver,shen2019networks, kim2021cyclemorph, liu2021rethinking}. Although, classic techniques have demonstrated potential results, a major bottleneck is the costly optimisation schemes needed for obtaining plausible results.
The second category is the focus of our interest in this work. In this section, we review the existing techniques.
%-------------------------------------------------------------------------
\subsection{Learning-based Techniques for Image Registration}
%Early works in the literature are based on classic registration frameworks, where different parametrisation and constraints were used. For example, B-spline curves~\cite{rueckert1999nonrigid}, momentum-parametrisation~\cite{vialard2012diffeomorphic}, constraints based spatial regularity~\cite{haber2007image} and stationary velocity fields\cite{beg2005computing,hart2009optimal}. Although, classic techniques have demonstrated potential results, a major bottleneck is the costly optimisation schemes needed for obtaining plausible results. This shortcoming prevents such frameworks to be feasible for handling a vast amount of data. With the aim of paving the way for more efficient computational solutions, learning-based image registration techniques have been recently proposed, where the majority of solutions can fall into either supervised or unsupervised techniques. @Angie: this paragraph has been deleted and added a few lines above.

A set of techniques have been proposed for supervised image registration, where convolutional neural networks (CNNs) are a de facto standard in the models; for example, the works of that~\cite{yang2017quicksilver,sokooti2017nonrigid,rohe2017svf,cao2017deformable,cao2018deformable}. Whilst supervised techniques have reported great performance, they require the ground truth deformation fields or segmentation masks. This requirement is particularly difficult in medical image registration. Existing techniques mitigate somehow that constraint by either relying on using classic techniques for getting a good ground truth estimation or using synthetic data. However, the registration performance is conditioned to the quality of the ground truth and/or the synthetic data pre-processing. 

To overcome the practical limitations of supervised techniques, 
another set of solutions has focused on designing unsupervised models. 
The authors of~\cite{krebs2018learning} use 
a statistical regularisation term to learn a low-dimensional stochastic parametrisation of the deformations. The works of that~\cite{de2017end,de2019deep} use B-spline parametrisation in a multi-stage framework, this technique enforces a coarse-to-fine learning process. Balakrishnan et al.~\cite{balakrishnan2018unsupervised,balakrishnan2019voxelmorph} introduced VoxelMorph, a cross-correlation CNN unsupervised framework that includes a spatial transform layer. A deep recursive cascade architecture was introduced in \cite{zhao2019recursive}, where a core of the model is a shared-weight cascading strategy. 

In more recent works, Mok et al.~\cite{mok2020fast} introduced a symmetric diffeomorphic framework called SYMNet, where authors guarantee topology preservation by introducing an orientation-consistent regulariser. The authors of~\cite{liu2020contrastive} proposed a contrastive registration architecture that fusions image-level registration and feature-level contrastive representation. CycleMorph was proposed in \cite{kim2021cyclemorph}, which uses cycle consistency. A key point of that work is that cycle consistency can provide a form of implicit regularisation for topology preservation. The authors of that~\cite{ye2021deeptag} introduced a bidirectional diffeomorphic network, that technique enforces topology-preservation and inevitability of the deformation. 

{\textbf{Connection with Image Segmentation.} There is an inherent connection between image registration and segmentation. One can use the computed mapping, between the unaligned and reference images, to project the segmentation mask of the reference image into the coordinate system of the unaligned image~\cite{liu2020contrastive,wang2020ltnet}. This process is called registration-based segmentation (aka atlas-based registration). In this work, we use this observation to unify the registration and segmentation process within one framework.}

\subsection{Vision Transformers \& Constrastive Learning}
A great focus of attention has been given to Vision Transformer (ViT)~\cite{dosovitskiy2020image} due to their performance-speed gain in classification tasks. After the work of Dosovitskiy et al.~\cite{dosovitskiy2020image}, several improvements have been proposed such as the techniques of~\cite{chu2021we, han2021transformer, wang2021pyramid} and for other tasks such as semantic segmentation and image detection~\cite{liu2021swin,wang2021pyramid}. However, there are only a few works that tackle the challenges of dealing with medical imaging tasks, where the focus has been mainly on segmentation e.g.\cite{chen2021transunet,hatamizadeh2021unetr}. 

\begin{figure*}[t]
 \centering
 \includegraphics[width=\textwidth]{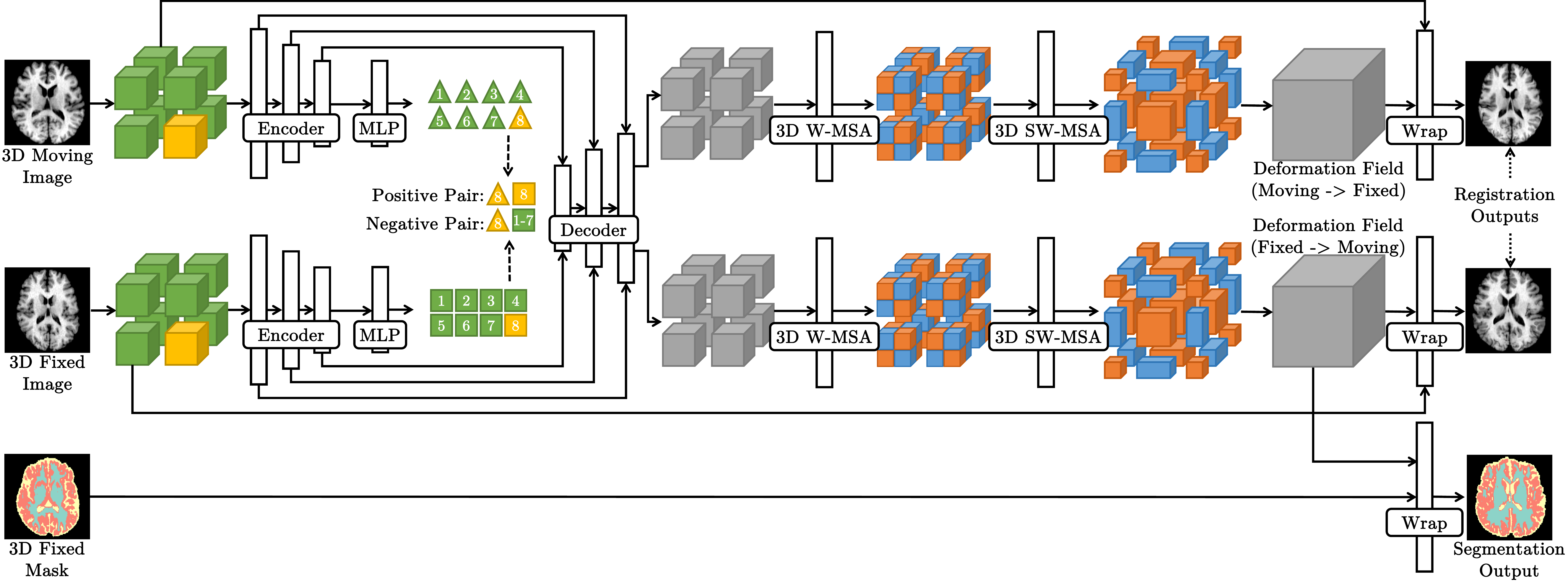} 
 \centering
 \caption{From left to right.{ Our PC-SwinMorph first generates non-overlap patches from the two input images, which are fed into two weight-shared CNN encoders. Followed by two MLP layers, the generated CNN features are projected to a latent space to obtain the patch representations (1-8 triangles and squares). {Detailed patchwise contrastive mechanism is shown in Fig.~\ref{fig:patchwise_contrastive}}. Based on patch representations, positive and negative pairs are sampled for patch-level contrastive learning (omitted from the figure). Then it recursively concatenates and enlarges the contrasted CNN feature with skip connection to reconstruct two sets of the deformation field patches.
 We use a 3D W-MSA and a 3D SW-MSA module to refine and stitch the deformation field patches to obtain the full deformation field.
 Using the full deformation fields, we wrap the moving image to the fixed image, and vice versa.
 After the training registration process, we also adopt the full deformation field to transfer the segmentation mask for fixed masks to obtain the segmentation mask of the moving image. All inputs and outputs are 3D volumes, and all the operations are implemented in 3D.}
 }
 \label{fig:networkTeaser}
\end{figure*}

By contrast, the works reported for medical image registration using transformers are scarce. The work in~\cite{chen2021vit} proposed a hybrid ConvNet-Transformer architecture for self-supervised volumetric image registration. The authors of~\cite{zhang2021learning} proposed a dual transformer network, where a self-attention scheme considers the inter- and intra- image context. These initial works showed the potential of vision transformers for image registration. In particular, the advantage is integrating more easily not only local but also global embeddings. However, there are still limitations on how self-attention schemes can better and more efficiently handle the correspondences between images.

Let us also mention the closely connected problem to our technique -- representation learning from the {unsupervised contrastive learning} perspective~\cite{hadsell2006dimensionality}. Contrastive learning techniques seek to learn similarities between sample pairs {without supervision}. Following this perspective, several techniques have been proposed including that of~\cite{chen2020simple, he2020momentum, tian2020contrastive}. Whilst the majority of existing techniques have been mainly applied for classification and segmentation tasks \cite{chen2020simple, he2020momentum,zhao2021contrastive}, the number of works reported for medical image registration is very limited. To the best of our knowledge, the work of Liu et al.~\cite{liu2020contrastive} is the only one reported for unsupervised registration and the closest to our purpose. In that work, the authors embedded a contrastive feature representation in the registration network to enforce feature maps with richer information.

%\textbf{Existing Techniques \& Comparison to our Work.} \Angie{we will add a few lines highlighting the differences wrt of existing works. This will be updated as the last thing}\Angie{to be updated}

\section{PC-SwinMorph: An Unsupervised Registration and Segmentation Framework} \label{sec:method}

In this section, we first introduce the necessary preliminaries for our technique. We then describe our proposed unified unsupervised framework, called PC-SwinMorph %\lihao{We should explain why this name when we first introduce this name.} @Angie: yes added in the intro
for registration and segmentation. We highlight two core strategies in our approach: i) patchwise contrastive learning, and ii) patches stitching using a shifted-window multi-head self-attention module. 

%the basic concepts and notations of a registration process, and then describe our proposed unified unsupervised joint registration and segmentation network. Then we describe the two ideas of our network design in detail: i) patchwise contrastive learning, and ii) patches stitching using shifted-window multi-head self-attention. 

%-------------------------------------------------------------------------
\subsection{Preliminaries \& Workflow Overview}
We first provide the essential preliminaries for our proposed framework.
%for image registration.
%Before the introduction of the network, we first introduce the basic concepts and notations for the registration process. 
%
Let $x$ and $y$ denote the moving (unaligned) and fixed (reference) 3D images respectively, where $x,$ $y \in \mathbb{Z}^{w \times h \times d}$. We refer to $w$, $h$, $d$ as the width, height, depth of the 3D images, where $\mathbb{Z}^{w \times h \times d} \subset \mathbb{Z}^3$. 
%\medskip
%Given two 3D images, we denote the moving image as $x$, and the fixed image as $y$, where $x$ and $y \in \mathbb{Z}^{w \times h \times d}$.
%
%In this notation, $w$, $h$, $d$ is the width, height, depth of the 3D images, where $\mathbb{Z}^{w \times h \times d} \subset \mathbb{Z}^3$.
%
We also denote the deformation field from $x$ to $y$ as $z_{x \rightarrow y}$, where $z_{x \rightarrow y} \in \mathbb{Z}^{w \times h \times d \times 3}$.
The deformation fields for 3D images are in a 4 dimensional space, i.e., $\mathbb{Z}^{w \times h \times d \times 3} \subset \mathbb{Z}^4$.
The four dimensions refer to each channel containing the pixel moving information in the $w$, $h$, $d$ axis, respectively.
%In the four dimensions, each channel contains the pixel moving information in the $w$, $h$, $d$ axis, respectively.
%
Moveover, the deformation field $z_{x \rightarrow y}$ is parametrised with a spatial transformation function denoted as $\psi_{z_{x \rightarrow y}}$, such that, the registered results $x \circ \psi_{z_{x \rightarrow y}}$ is aligned with fixed image $y$.

The workflow of our PC-SwinMorph framework is displayed in Figure~\ref{fig:networkTeaser}, which unifies unsupervised registration and segmentation tasks. 
%Figure~\ref{fig:introduction_network} shows our proposed VoxelMorph-based unified unsupervised joint registration and segmentation architecture.
%
Our framework uses an encoder to extract the CNN feature maps from the two given 3D images $x$~and~$y$.
We then seek to estimate two deformation fields $z_{x \rightarrow y}$ and $z_{y \rightarrow x}$ from the extracted CNN feature maps with skip connections.
After we obtain the deformation fields, we perform registration by using a spatial transformer\cite{jaderberg2016spatial} to warp the moving image $x$ and the deformation field $z_{x \rightarrow y}$. We can then obtain the registration output ($x \circ \psi_{z_{x \rightarrow y}}$), where~$\circ$ denotes the wrap operation.
Similarly, we can also use a spatial transformer to wrap the fixed image $y$ and the deformation field $z_{y \rightarrow z}$ to obtain $y \circ \psi_{z_{y \rightarrow x}}$.

{After registration, we also use the spatial transformer 
%Finally, based on the deformation field $z_{y \rightarrow z}$, 
to wrap deformation field $z_{y \rightarrow z}$ and the segmentation mask of the image $y$. By doing this, we can obtain
%to obtain 
the segmentation mask for any image $x$.}
We underline that 
%Since 
there is no segmentation mask used in the training registration process, and the segmentation mask is only used in the testing segmentation stage. Therefore, our framework 
%claim that this
is a unified unsupervised registration and segmentation network.

\subsection{{Patch-based Strategy I:} Patchwise Contrastive Registration}
A core of our technique is a patchwise strategy for image registration. It enforces an efficient and accurate registration output. We provide details next.
%In order to carry out an easier but more effective registration, we proposed to perform patchwise registration.

\textbf{Patch Generation.} 
We first generate patches from the two 3D images instead of feeding directly the moving image $x$ and the fixed image $y$ as input.
%Instead of taking the moving image $x$ and the fixed image $y$ as input directly, we first generate patches from the two 3D images.
%
To do this, we evenly partition the 3D images into $n^3$ patches, that is, $n$ pieces along the $w$, $h$, $d$ axis without overlapping.
%
%We split the 3D images into $n^3$ patches by evenly partitioning them into $n$ pieces along the $w$, $h$, $d$ axis without overlapping.
%
We denote the generated patches from the moving and fixed images as $p^x_i$ and $p^y_i$, where $i \in [1, n]$ and $p^x_i$ and $p^y_i \in \mathbb{Z}^{\frac{w}{n} \times \frac{h}{n} \times \frac{d}{n}} $.
For example, as shown in the Figure~\ref{fig:networkTeaser}, we set $n$ as 2. 
Therefore, both moving images and fixed images are split into $2^3$ patches (8 patches).

\begin{figure*}[t]
\centering
\includegraphics[width=0.8\textwidth]{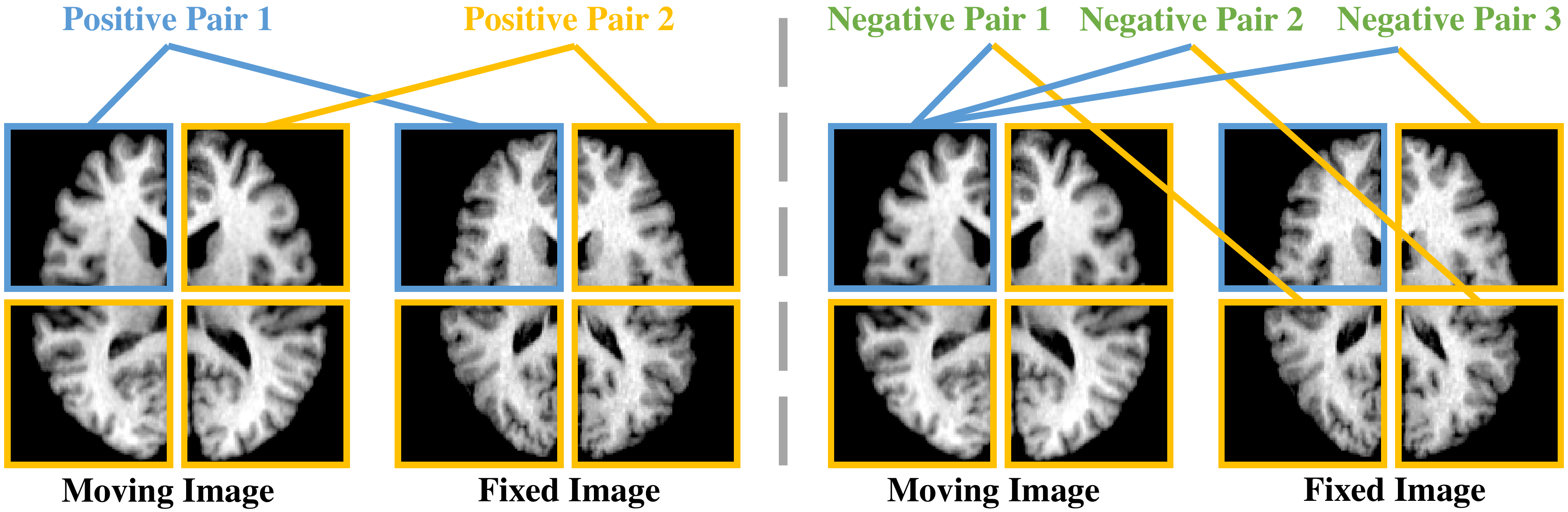} 
\centering
\caption{Patchwise contrastive mechanism. (Left side) Positive pairs are sampled from same-position patches in difference images, for example, the top-left patches of the moving and fixed image are \textcolor{dodgerblue}{\textbf{positive pair - 1}}, and the top-right patches of the moving and fixed image are \textcolor{selectiveyellow}{\textbf{positive pair - 2}}. (Right side) Any pairs from difference position are sampled as negative pairs, such as the the top-left moving patch and the bottom-left fixed patch are \textcolor{lightgreen}{\textbf{negative pair - 1}}, and top-left moving patch and the top-right fixed patch are \textcolor{lightgreen}{\textbf{negative pair - 2}}. For visualisation purposes, the 3D volumes are represented as 2D slices.}
\label{fig:patchwise_contrastive}
\end{figure*}

\textbf{Patchwise Registration.} We first select a patch pair from the moving and fixed partitioned images at the same position ($p^x_i$ and $p^y_i$).
%patch with the same cutting position ($p^x_i$ and $p^y_i$)
%We first select a pair of the moving and fixed patches with the same cutting position ($p^x_i$ and $p^y_i$), 
We then feed them into a two-symmetric weight-shared 3D encoder to extract the highly semantic CNN feature maps (see the two encoders in Figure~\ref{fig:networkTeaser}). 
We use a single decoder to integrate all the CNN feature maps generated by the two encoders.
Specifically, we recursively combine the CNN feature maps from high-to-low level (low-to-high image resolutions)
%level (low resolution) to low-level (high resolutions), 
%until we 
to reconstruct two deformation field patches ($p^{z_{x \rightarrow y}}_i$ and $p^{z_{y \rightarrow x}}_i$) that have the same resolution as the input patches ($p^x_i$ and $p^y_i$). See the decoder part in Figure~\ref{fig:networkTeaser}.
We repeat this process until every same-position patch pair has been fed through the encoder-decoder architecture to produce their corresponding two deformation field patches. 

After obtaining the patchwise deformation fields, we stitch them to produce the full deformation field ($\hat{z}_{x \rightarrow y}$ and $\hat{z}_{y \rightarrow x}$). To achieve this, we use a 3D Swin Transformer Block~\cite{liu2021swin}. See the 3D W-MSA and 3D SW-MSA modules in Figure~\ref{fig:networkTeaser}.
%
%We describe the stitching process in the next section.
%
We then use a spatial transformer to warp the moving image $x$ and the stitched deformation field $\hat{z}_{x \rightarrow y}$ to 
%generate the registered result 
obtain the composition $x \circ \psi_{\hat{z}_{x \rightarrow y}}$, and do the inverse registration from $y$ to $x$.
%
%Finally, 

As part of our unified framework for unsupervised registration and segmentation, we can now use the estimated deformation field to do segmentation tasks. {More precisely, in the testing stage, we use a spatial transformer, to warp the segmentation mask of the fixed image $y_{seg}$ and (stitched) deformation field $\hat{z}_{y \rightarrow x}$, to generate the segmentation result ($y_{seg} \circ \psi_{\hat{z}_{y \rightarrow x}}$).}

\smallskip
\textbf{Reconstruction \& Regularisation Terms.} 
%In the training process, 
During the training process, we use two losses to guarantee a robust registration process.
We first use a reconstruction loss, 
%The first loss is reconstruction loss,
which %is designed to make 
enforces a plausible mapping to get the {registered results} as closest as possible to the {template images}. 
%the registered result as similar as possible to its template.
%
In our work, we use a normalised local cross-correlation loss (NCC loss) as the reconstruction loss.
We denote the two input images as $x$ and $y$. 
Then the local mean of $x$ and $y$ at pixel $p$ are denoted as $\bar{x}(p)$ and $\bar{y}(p)$, respectively. 
The NCC loss is given as follow:
\begin{equation}
\begin{aligned} \label{loss_ncc}
  \mathcal{L}_{ncc}(x, y) = 
 & \sum_{p \in \Omega} \frac{\sum_{p_i} (x(p_i) - \bar{x}(p)) \cdot (y(p_i) - \bar{y}(p))}{\sqrt{\sum_{p_i} (x(p_i) - \bar{x}(p))^2 \cdot \sum_{p_i} (y(p_i) - \bar{y}(p))^2}}, 
\end{aligned}
\end{equation}
where the local mean $\bar{x}(p)$ and $\bar{y}(p)$ are calculated over a local window centered at pixel $p$ with window length of 9, and in the domain $\Omega \subset \mathbb{Z}^{w \times h \times d}$.
Our reconstruction loss reads:
\begin{equation} \label{loss_recon}
 \mathcal{L}_{recon} = \mathcal{L}_{ncc}(x \circ \psi_{\hat{z}_{x \rightarrow y}}, y) + \mathcal{L}_{ncc}(y \circ \psi_{\hat{z}_{y \rightarrow x}}, x). 
\end{equation}

We also include %a smooth term. We
%The second loss is a smooth loss, use
an L2 diffusion regulariser on the spatial gradients to obtain a smooth deformation field. It reads:
%as smooth as possible:
\begin{equation} \label{loss_smooth}
 \mathcal{L}_{smooth} = \sum_{p \in \Omega} ||\nabla \psi_{\hat{z}_{y \rightarrow x}}||^2 + \sum_{p \in \Omega} ||\nabla \psi_{\hat{z}_{y \rightarrow x}}||^2.
\end{equation}

\smallskip
\textbf{Patchwise Contrastive Loss.} %In unsupervised representation learning tasks,
In representation learning, contrastive learning has been a successful perspective to learn %representation
distinctiveness. 
The main idea of contrastive learning is to maximise the similarity between images and their augmented views, whilst minimising the similarity between images from different groups.
In contrast to existing contrastive methods, which select the positive and negative pairs between images in the dataset, we proposed to select the positive and negative pairs within the image internally; {as shown in Fig.~\ref{fig:patchwise_contrastive}}.
%Different from contrastive methods which select the positive and negative pairs between images in the dataset, we proposed to select the positive and negative pairs from within the image internally.
%
More precisely, after we recursively fed the moving and fixed patches $p^x_i$ and $p^y_i$ into the CNN encoders, we can obtain a set of high-level semantic CNN features maps for each patch.
We use two linear projection layers {(see the MLP in Figure~\ref{fig:networkTeaser})} to map the high-level CNN semantic feature maps, for the moving and fixed patches, to a latent space, separately.
Hence, the projected features are a calculated representation of the moving patch and fixed patch.
We denote the projected features as $f^x_i$ and $f^y_i$ for the moving and fixed patch respectively, {see the triangles and squares tagged as 1 to 8 in Figure~\ref{fig:networkTeaser}.}
Between the two sets of projected features, we consider as a positive pair any part from the same partition position ($f^x_i$ and $f^y_i$), otherwise a negative pair ($f^x_i$ and $f^y_j$, where $i \neq j$).
%we consider that any pair from the same partition position is a positive pair ($f^x_i$ and $f^y_i$), otherwise, a 
%whereas any pair with a different partition position is a negative pair ($f^x_i$ and $f^y_j$, where $i \neq j$).
% %
% Due to the similarity between the medical images (i.e., the basic biological anatomy of medical images), there is no biological difference between the same regions from different images.
% %
% For example, the centre region of the brain images is a sub-cortical organ called the hippocampus.
%
%
We then consider the following contrastive loss from $f^x_i$ to $f^y_i$ expressed as:

\begin{equation}
\begin{aligned} \label{loss_patch_contrast_1}
 \mathcal{L}^{i}_{contrast}(f^x_i, f^y_i) = - \log \frac{\exp(sim(f^x_i, f^y_i) / \tau)}{\sum^{n}_{j=1} \exp(sim(f^x_i, f^y_j) / \tau)},
\end{aligned}
\end{equation}
where $sim(u, v) = \frac{u^\mathrm{T}v}{||u||\cdot||v||}$ is the cosine similarity between $u$ and $v$. Moreover, $\tau$ is a temperature hyperparameter set as 1, and $n$ is the number of patches.
Similarily, the the contrastive loss from $f^y_i$ to $f^x_i$ which reads:

\begin{equation}
\begin{aligned} \label{loss_patch_contrast_2}
 \mathcal{L}^{i}_{contrast}(f^y_i, f^x_i) = - \log \frac{\exp(sim(f^y_i, f^x_i) / \tau)}{\sum^{n}_{j=1} \exp(sim(f^y_i, f^x_j) / \tau)}.
\end{aligned}
\end{equation}
Hence, the final patchwise contrastive loss is calculated as a weighted sum of~\eqref{loss_patch_contrast_1} and~\eqref{loss_patch_contrast_2} as:
%previous two losses:
{\small
\begin{equation}
\begin{aligned} \label{loss_patch_contrast_final}
 \mathcal{L}_{contrast} = 
 & - \frac{1}{2n} \sum^{n}_{i=1} (\mathcal{L}^{i}_{contrast}(f^x_i, f^y_i) + \mathcal{L}^{i}_{contrast}(f^y_i, f^x_i)).
\end{aligned}
\end{equation}
}

\subsection{{Patch-based Strategy II:} Patches Stitching with Shifted-window Multi-head Self-attention}

\begin{figure}[t!]
 \centering
 \includegraphics[width=1\linewidth]{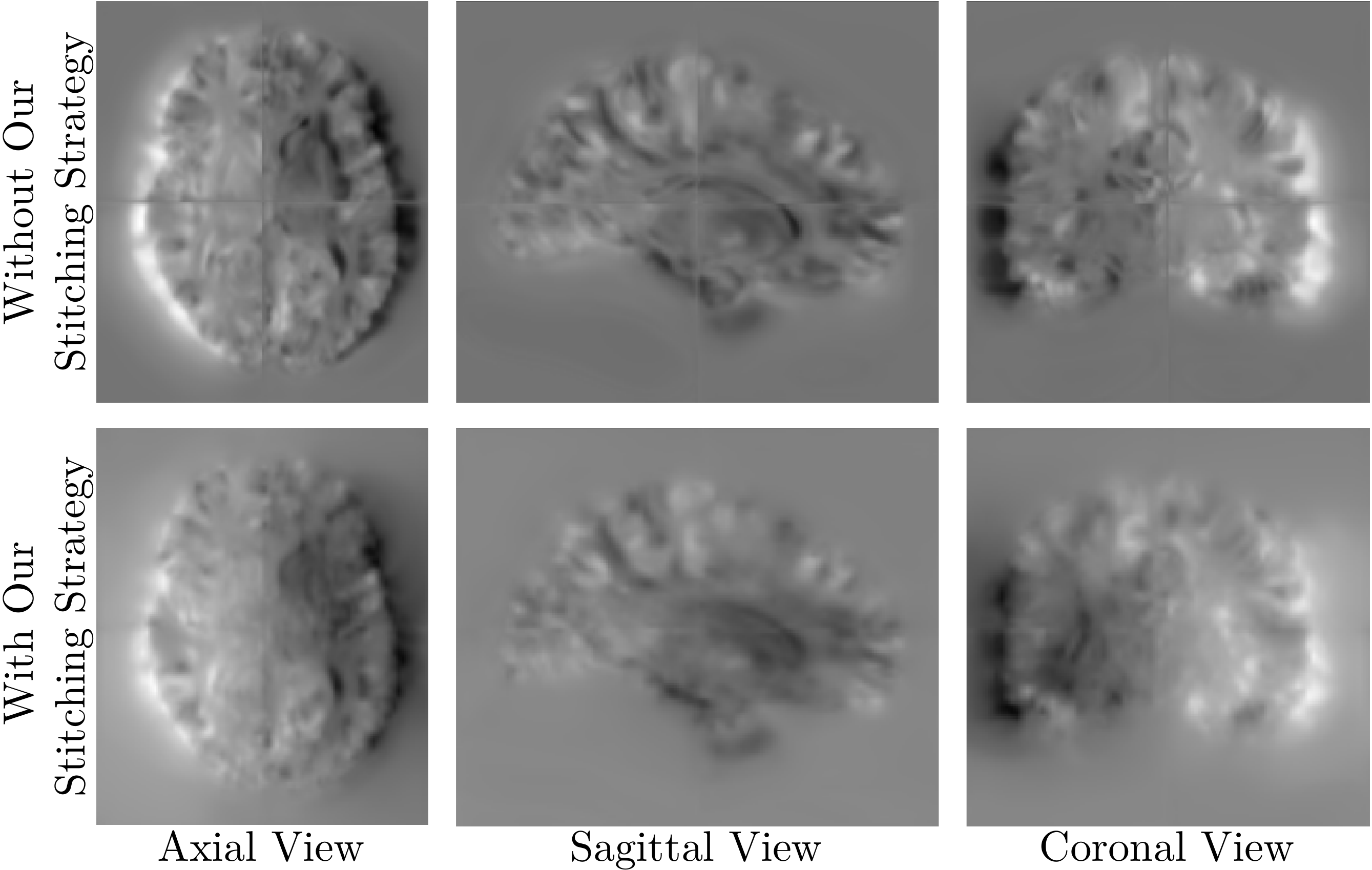}
 \caption{{Deformation field learned with and without the proposed stitching strategy. Three columns show the sample slice of the 3D deformation field from different views. Direct patch stitching (first row) yields results with clear discontinues between patches, whilst stitching with the proposed multi-head self-attention (second row) produces smooth and continuous deformation fields with the gap between patches nearly invisible.}}
 \label{stitching_effect}
\end{figure}

The multi-head self-attention (MSA) module~\cite{vaswani2017attention} has been proved to be an effective tool for capturing content relations from within the image. 
Unlike CNN-based networks that use a hierarchical convolutional layer to expand the reception field from local to global, the MSA module can directly calculate the similarity of all non-overlapping patches within the image.
%
%The detailed description of MSA can be found in ~\ref{}. already cited a few lines above
%
In our work, to alleviate the blurring effect from the process of direct patch stitching ({see the top row of Figure~\ref{stitching_effect}}), we propose to stitch the patches with multi-head self-attention.
{Specifically and for computational efficiency, we use the improved 3D window/shifted-window multi-head self-attention (3D W-MSA and 3D SW-MSA) from Swin Transformer~\cite{liu2021swin}.}
% to explore the relations between different non-overlapping deformation field patches.
% We recap that our goal is to reconstruct a refined $\hat{z}_{x \rightarrow y}$ and $\hat{z}_{y \rightarrow x}$ from the two set of deformation field patches $p^{z_{x \rightarrow y}}_i$ and $p^{z_{y \rightarrow z}}_i$ ($i \in n$), respectively.
%

\textbf{3D W-MSA.} To align with the definition of Swin Transformer, we define each deformation field patch $p^{z_{x \rightarrow y}}_i$ as a window.
Each window is further splited evenly into $m \times m \times m $ small non-overlapping regions $r^{z_{x \rightarrow y}}_{ij}$ where $i \in n^3, i \in m^3$.
The original MSA performs the computation directly on the regions of an image with a size of $w \times h \times d$.
Whereas the W-MSA calculates region relations inside windows, which can largely save computation time.
The computational complexity of the two modules are listed as follow:
\begin{equation}
\begin{aligned} \label{complextity_msa}
 \Omega(MSA) & = 4 \cdot whd \cdot C^2 + 2 \cdot (whd)^2 \cdot C \\
 \Omega(W-MSA) & = 4 \cdot whd \cdot C^2 + 2 \cdot m^3 \cdot (whd) \cdot C,
\end{aligned}
\end{equation}
where the time complexity for the MSA module is quadratic to volume size $whd$, the time complexity for the W-MSA module is linear to volume size where $m$ is set as 4, and C is the {image spatial channel number} 3 in our experiment. Hence, with the W-WSA module, the computational time is fast, especially apply to 3D medical images.

\textbf{3D SW-MSA.} \textit{One core disadvantage of 3D W-MSA is that it lacks information exchange between windows}, since all the computation is performed on regions within a window, {which means that simply applying the 3D W-MSA on the deformation field patches is not enough for dealing with the stitching effect see Figure~\ref{stitching_effect}.
%cannot erase the stitching effect.
%
Hence, based on the 3D W-MSA outputs, we further use the 3D SW-MSA to enhance information change between windows for stitching effect alleviating.}
We follow the cyclic-shifting strategy and move the window along the diagonal direction by one region \big($\frac{w/n}{m} \times \frac{d/n}{m} \times \frac{h/n}{m}$\big).
By cyclic-shifting, the generated window goes from $n^3$ to $(n+1)^3$.
As shown in Figure~\ref{fig:networkTeaser}, after the 3D SW-MSA, the window number increased from 8 to 27.
{
With the increase of windows, regions from different windows are mingled %and mixed @Angie: I remove mixed as both words refer to the same
for calculation, which allows information exchanges to erase the stitching effect.
After the two modules, we stitch the self-attended window to obtain the stitched deformation field.
We underline that our stitching strategy is like a clip-on function to refine the deformation field without introducing an additional loss term.
Whilst the detailed description of the architecture of W-MSA and SW-MSA can be found in~\cite{liu2021swin}.}

% where the window in the 8 corner is smaller ($\frac{w/n}{m} \times \frac{d/n}{m} \times \frac{h/n}{m}$), and window in the center is the same size as previous step ($\frac{w}{n} \times \frac{d}{n} \times \frac{h}{n}$).
% %
% For computational simplicity, we pad the small windows to the larger window size and then do the calculation with \Angieb{mask}.
% %
% We mask out the padded region, therefore, only the original windowed region is calculated in an efficient batch computational approach.
% %
% After the two modules, we reverse the window back and stitch the attended window to obtain the stitched deformation field.

%\clearpage

\section{Experimental Results}
\label{sec:experiment}
In this section, we detail the set of experiments performed to evaluate our proposed unified 
%unsupervised registration and segmentation 
framework.

\begin{figure*}[t]
 \centering
 \includegraphics[width=\textwidth]{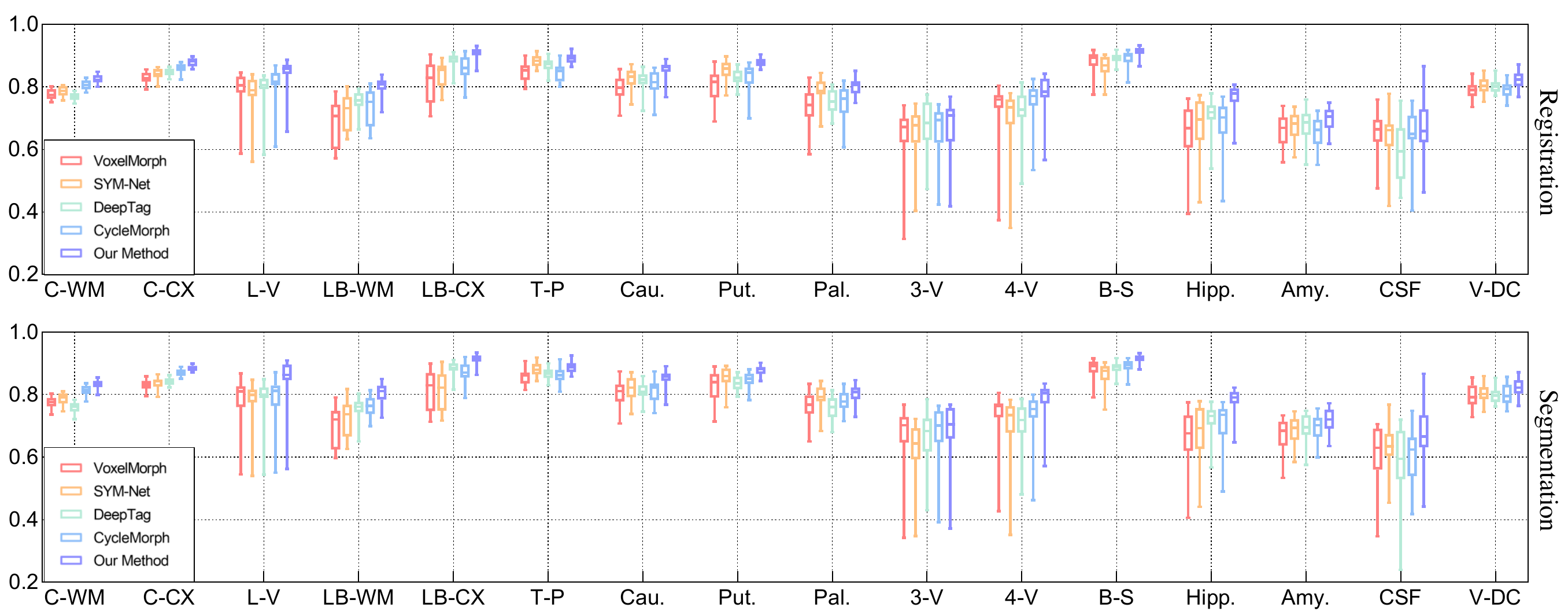} 
 \centering
 \caption{{Boxplot comparisons between our proposed PC-SwinMorph technique and SOTA techniques on the Candi dataset. The Dice similarity metric is reported per each region in the brain. The numerics per region can be found in supplementary material.}}
 \label{candi_results}
\end{figure*}

\begin{figure}[t]
 \centering
 \includegraphics[width=\linewidth]{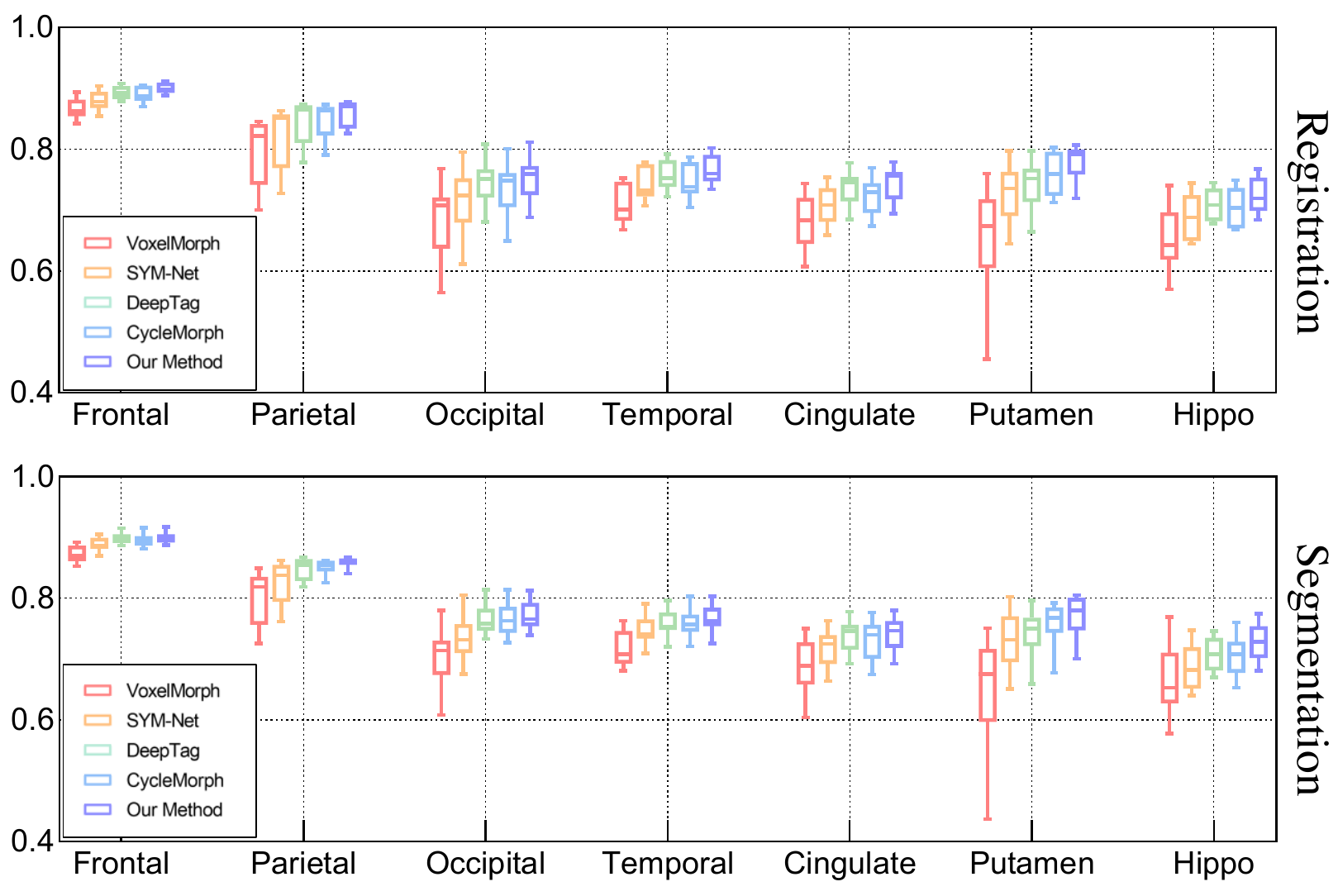} 
 \centering
 \caption{{Boxplot comparisons between our proposed PC-SwinMorph technique and SOTA techniques on the LPBA40 dataset. The Dice similarity metric is displayed  per each region in the brain. We reported the exact values of each technique in the supplementary material.}}
 \label{lpba40_results}
\end{figure}

% \input{PC-SwinMorph/table/table_1}
% \input{PC-SwinMorph/table/table_2}

%-------------------------------------------------------------------------
\subsection{Datasets Description}
We evaluate our framework using %We use 
two publicly available datasets: %to evaluate our technique:
the Child and Adolescent Neuro Development Initiative (CANDI) dataset\cite{722211} and the LONI Probabilistic Brain Atlas (LPBA40) dataset\cite{shattuck_construction_2008}.

\textbf{CANDI Dataset.} CANDI dataset is comprised of 103 T1-weighted MRI scans with anatomic segmentation labels. The volume size of the MRI scans ranges from $256 \times~256~\times~128$ to $256 \times 256 \times 158$ voxels with a uniform space of $ 0.9375\times 0.9375 \times 1.5$ $mm^3$. We followed the convention used in ~\cite{wang2020ltnet} to group the same components from left and right brains together, forming overall 16 anatomical regions.
%16 segmentation labels. 
%To improve the 
For computational efficiency, we crop the volume to $160 \times 160 \times 128$ around the centre of the brain, which is large enough to incorporate the whole brain region. 

\textbf{LPBA40 Dataset.} LPBA40 dataset contains 40 T1-weight 3D brain volumes from 40 healthy humans. The size of 3D volumes is $181 \times 217 \times 108$ with a uniform space of $1 \times 1 \times 1$ $mm^3$. The 3D brain volume was manually segmented to identify 56 structures. Similar to the CANDI dataset, we crop the data to $160 \times 192 \times 160$ around the centre of 3D volumes to reduce the size of the volume whilst preserving all the brain regions. All the 56 structures are grouped into seven large regions in order to display the segmentation results more intuitively~\cite{liu2020contrastive}.

\subsection{Implementation Details} \label{implementationDetails}
%We now describe the implementation details along with the training and testing scheme 
%that we 
%followed to produce the reported results.

\textbf{Data Pre-processing.} 
%We follow a standard pre-processing protocol for medical images. 
We normalise the volumes to have zero mean and unit variance. 
% Furthermore, we resample from the original cropped volumes to reduce the size of the volumes to half at all axis. 
{For the CANDI dataset, we follow the data splitting in~\cite{wang2020ltnet} and select 20 volumes as test data, 1 volume as the reference image, and the rest as training data. 
For the LPBA40 dataset, we set the first volume as the reference image, the next 29 images as training images, and the last 10 images as testing images.}

\textbf{Training Scheme.}
%In the training scheme,
During the training stage, the parameters of all convolutional layers are initialised by following the initialisation protocol of \cite{he2015delving}. 
% The parameters of all batch normalization layers are also initialised by drawing random variables from Gaussian distribution.
Adam optimiser is used during training with the initial learning rate setting as $10^{-3}$. The learning rate decays by 0.1 scale every 50 epochs and terminates after 200 epochs. The batch size for both datasets is 1. All models are run on an NVIDIA A100 GPU with 80G RAM, which takes around 6 hours to train the model on the CANDI dataset and around 4 hours on the LPBA40 dataset.

\begin{table}[t]
    \centering
    \resizebox{0.85\linewidth}{!}{
    \begin{tabular}{lcccc} \hline \toprule[1pt]
    & \multicolumn{2}{c}{\textbf{\textsc{CANDI}}}
    & \multicolumn{2}{c}{\textbf{\textsc{LPBA40}}} \\
    \midrule[0.8pt] 
    
    {} & \cellcolor[HTML]{EFEFEF} \textit{Reg.} & \cellcolor[HTML]{EFEFEF} \textit{Seg.} & \cellcolor[HTML]{EFEFEF} \textit{Reg.} & \cellcolor[HTML]{EFEFEF} \textit{Seg.}
    \\
    \midrule[0.8pt]
     VoxelMorph    & 0.753  & 0.759   & 0.720  & 0.728     \\
     SYMNet        & 0.769  & 0.760   & 0.755  & 0.762     \\
     DeepTag       & 0.773  & 0.771   & 0.775  & 0.780     \\
     CycleMorph    & 0.772  & 0.780   & 0.772  & 0.782     \\ \midrule[0.8pt]
     PC-SwinMorph  &\textbf{0.812}  & \textbf{0.817} & \textbf{0.791}  & \textbf{0.794}  \\ \midrule[0.8pt]
    \end{tabular}
    }
    \caption{Numerical comparision between our PC-SwinMorph technique and other state-of-the-arts methods. We denote  Registration as 'Reg'  and Segmentation as 'Seg' Measures averaged over all regions using the Dice coefficient.  Results on both CANDI and LPBA40 datasets for registration and segmentation tasks are presented, with the best performance highlighted in bold font.} 
    \label{avg_results}
    \end{table}

\textbf{Testing Scheme.}
During the testing process, the unsupervised segmentation is performed based on the learned parameters. We firstly fed the trained network with the moving image $x$ and fixed images $y$ to obtain the deformation field $\hat{z}_{y \rightarrow x}$. We then used a spatial transform network \cite{jaderberg2016spatial} to wrap the segmentation mask of fixed image ($y_{seg}$) and the deformation field $\hat{z}_{y \rightarrow x}$, we then obtain the segmentation results of the moving image ($y_{seg} \circ \psi_{\hat{z}_{y \rightarrow x}}$)). With a single GPU (NVIDIA A100 GPU), our method can process around 3.2 brain images per second. 

\textbf{Evaluation Protocol.} %For a fair comparison, 
{To make our results comparable to other state-of-the-art methods,} %"For a fair comparison is repeated again in the later expression" @Angie: yes!
we use the $Dice$ similarity coefficient to evaluate the segmentation {and registration} quality of our model, which measures the overlap between ground truth masks and predicted segmentation results. \textit{The code with detailed description will be available with the publication.}

\begin{figure*}[t!]
  \centering
  \includegraphics[width=1\textwidth]{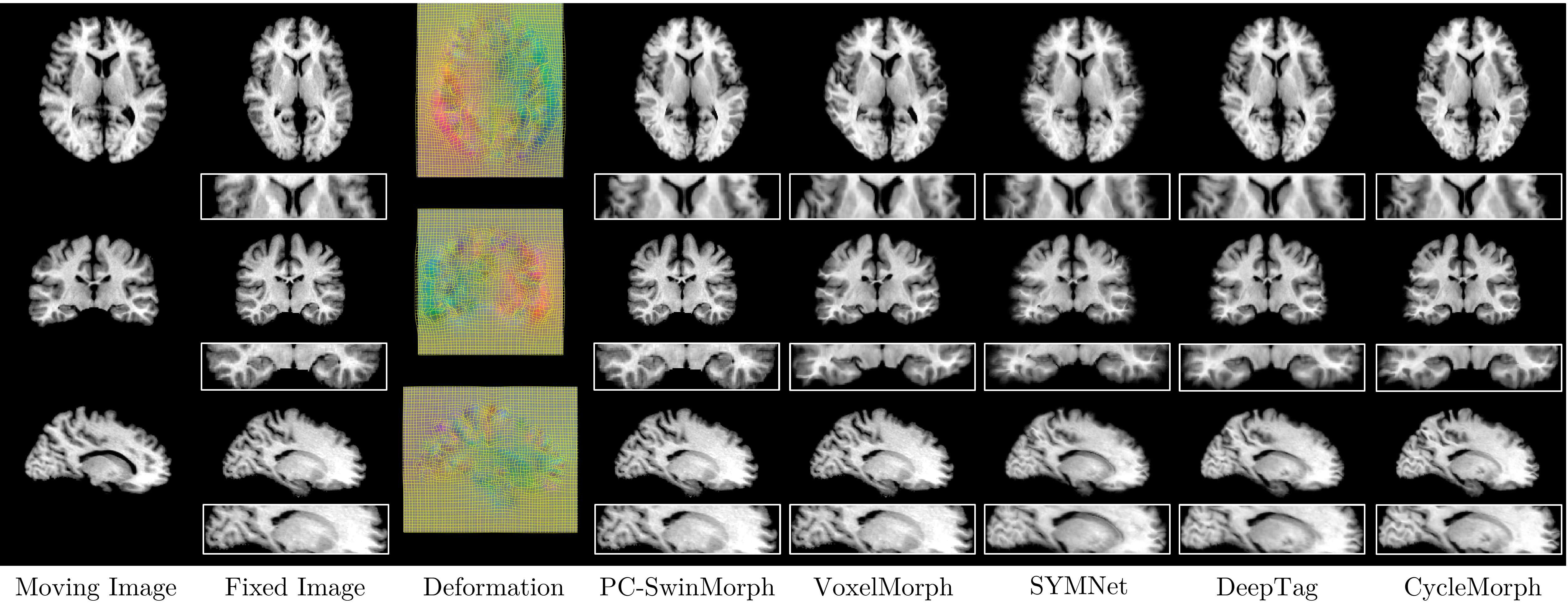}
  \caption{Visual comparisons between the proposed PC-SwinMorph technique, and other unsupervised SOTA techniques for registration. The rows show the three views from 3D volumes. The first two columns display the moving and fixed images, while columns 4-8 present the aligned images produced by PC-SwinMorph and other unsupervised SOTA techniques. The zoom-in views highlight regions that demonstrates the improvement of our method in terms of preserving the global brain structures and fine local details. The third column presents the deformation field computed by our method.}
  \label{visual_reg_result}
\end{figure*}

\begin{figure*}[t]
  \centering
  \includegraphics[width=0.875\textwidth]{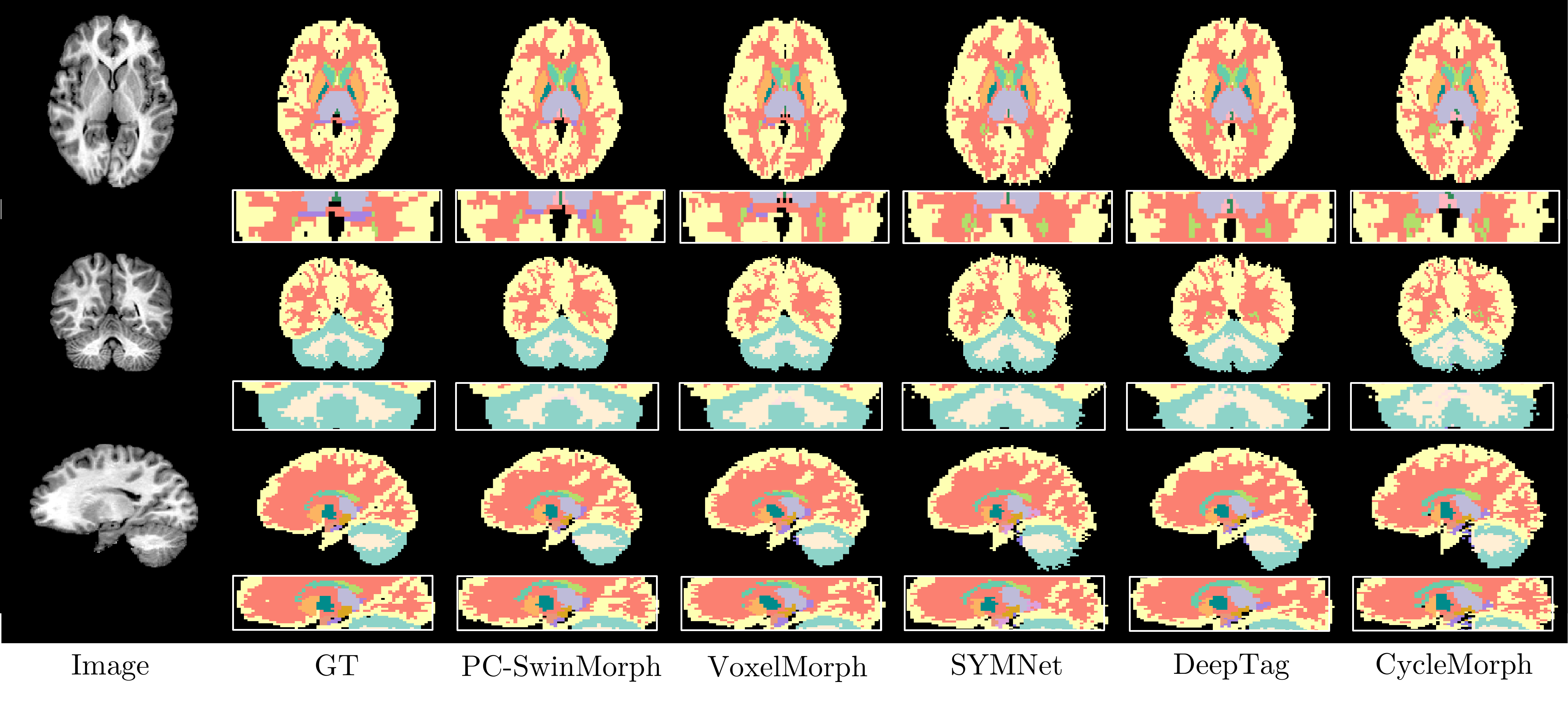}
  \caption{Visual comparisons between the proposed PC-SwinMorph technique, and other unsupervised SOTA techniques for segmentation. The rows show the three views from 3D volumes. The second column displays the ground truth (GT) results whilst columns 3-6 display predicted segmentation from PC-SwinMorph and other unsupervised SOTA techniques. Zoom-in views demonstrate that the proposed method preserved more details in different regions thus producing better segmentation results.}
  \label{visual_seg_result}
\end{figure*}

\subsection{Comparison to the State-of-the-Art Techniques} 
\label{sotaresults}
We compared our technique with four recent unsupervised brain segmentation methods, including VoxelMorph\cite{balakrishnan2019voxelmorph}, DeepTag\cite{ye2021deeptag}, SYMNet\cite{mok2020fast}, CycleMorph\cite{kim2021cyclemorph}.
In general, all methods are based on the fundamental architecture of VoxelMorph. 
{
%Among them, CycleMorph implemented cycle consistency architecture to enhance the deformation fields.
%
%While DeepTag and SYMNet modified the architecture by introducing bi-directional forward-backward learning and symmetric diffeomorphic framework.
For a fair comparison, all models use the same backbone architecture, VoxelMorph, which has been fine-tuned to achieve optimal performance.
%
% For the compared techniques, we use either the code provided by the authors or %Different models are reproduced by either reusing the public code provided by the authors or 
% following the architecture described in the original paper. Recommended parameters are used in all models.
}

\textbf{Numerical Comparison.}
%We start by comparing performance-wise our technique and existing techniques.
Table~\ref{avg_results}, Fig.~\ref{candi_results} and Fig.~\ref{lpba40_results} summarise performance-wise, in terms of the $Dice$ coefficient, the compared SOTA methods, and our PC-SwinMorph. 
The numbers are obtained using the testing scheme described in section~\ref{implementationDetails}. %above. 
In a closer look at the tables, we observe that for both data, our method outperforms all other SOTA methods by a large margin, including the overall performance as well as local performance in major regions. 
%
% Figure~\ref{candi_results} and \ref{lpba40_results} display the detailed results for all regions.
Particularly, {on the CANDI dataset} our results report an improvement of 5.9\% compared to VoxelMorph, and 3.9-4.3\% against the other compared techniques. 
%In general, registration dice and segmentation dice are around 5-6\% higher than the baseline VoxelMorph and at least 3-4\% higher than other SOTA methods. 
This performance gain is consistent on the LPBA40 dataset, where our proposed technique is {1.2-7.0\%} higher in performance than the SOTA methods.

\textbf{Visual Comparison.}
We support the numerical results with additional visual results 
% on the CANDI dataset %by other methods and the compared SOTA methods. 
for our technique and the compared ones.
Figure~\ref{visual_seg_result} shows some sample slices of the segmentation results predicted by VoxelMorph, SYMNet, DeepTag, CycleMorph, and our proposed method PC-SwinMorph.
Whilst the results produced by the compared SOTA techniques
%We observed that results predicted by all SOTA methods 
are anatomically meaningful, they fail to capture fine details in several regions. 
By contrast, PC-SwinMorph is able not only to produce a better output but also to capture details. 
%PC-SWinmorph captures more details in regions where other methods failed to do. 
The zoom-in views in Figure~\ref{visual_seg_result} highlights these effects.
Overall, PC-SwinMorph can better accommodate with fine details of the brain structure, producing segmentation closer to the ground truth. 
Figure~\ref{visual_reg_result} shows the wrapped images produced by different SOTA techniques as well as the proposed PC-SwinMorph. We can observe that our registration outputs are closer to the reference image, displaying fewer splitting effects from the patch generation whilst keeping fine details. \nopagebreak

\nopagebreak
\subsection{Ablation Study} We provide a set of experiments to further support the design of our technique.

\textbf{Contrastive Representation.} A contrastive feature learning mechanism is embedded into the registration architecture which promotes feature-level learning. The contrastive loss forces the network to contrast the difference between the two extracted CNN feature maps and therefore, the network is more discriminative to different images via contrasting unaligned images and reference images. As shown in Table \ref{ablation_study_results}, \textit{the testing results demonstrate this new mechanism significantly improves the segmentation performance by around 2-3\% upon the baseline model.}

\textbf{Patchwise Contrastive Learning.}
Based on the idea of contrastive learning, we introduce a patchwise contrastive learning strategy. It uses a
%introducing the patchwise contrastive learning using a
multilayer patch-based approach rather than operating on entire images. The Patchwise contrastive loss introduced in~\eqref{loss_patch_contrast_final}
%\(\mathcal{L}^{i}_{contrast}(f^x_i, f^y_i)\) 
encourages two corresponding patches, in the image, to map to a point in a learned feature space, {at the same time drawing negative if they match to other patches.} This mechanism further boosts the registration performance and produces better segmentation results. From the $Dice$ coefficient comparison reported in Table \ref{ablation_study_results}, we can observe that the patchwise contrastive approach offers an additional 2\% improvement with respect to the contrastive learning approach. This improvement is observed in both the registration and segmentation performance. This introduced strategy offers an overall improvement of 4\% when compared to the baseline model.
%upon the baseline model to be around 4\%.

\textbf{The effect of 3D W-MSA \& 3D SW-MSA.}
One of the main drawbacks of patchwise learning is the lack of information exchange between patches. The introduction of 3D W-MSA and 3D SW-MSA, which stitches the patches and enhances the performance across patches, has been demonstrated to be greatly effective. 
%its effectiveness convincingly in this study. 
As shown in Table \ref{ablation_study_results}, the segmentation and registration performance on the CANDI dataset has been improved by 2.65\% and 3.06\% respectively when using 3D SW-MSA.
%, improving the overall dice value to 0.812 and 0.817 for registration and segmentation respectively. 
This performance behavior is prevalent on LPBA40  demonstrating consistent performance.%\clearpage

\begin{table}[t]
\centering
\resizebox{0.92\linewidth}{!}{
\begin{tabular}{lcccc} \hline \toprule[1pt]
& \multicolumn{2}{c}{\textbf{\textsc{CANDI}}}
& \multicolumn{2}{c}{\textbf{\textsc{LPBA40}}} \\
\midrule[0.8pt] 

{} & \cellcolor[HTML]{EFEFEF} \textit{Reg.} & \cellcolor[HTML]{EFEFEF} \textit{Seg.} & \cellcolor[HTML]{EFEFEF} \textit{Reg.} & \cellcolor[HTML]{EFEFEF} \textit{Seg.}
\\
\midrule[0.8pt]
 {Baseline (B)}  & 0.753  & 0.759 & 0.720  & 0.728     \\
 {B + PW} & 0.779  & 0.777   & 0.751  & 0.762 \\
 {B + CL} & 0.771  & 0.772   & 0.760  & 0.766 \\
 {B + PW + CL} & 0.791  & 0.792   & 0.779  & 0.783 \\ \midrule[0.8pt]
 {B + PW + CL + SW} &\textbf{0.812}  & \textbf{0.817} & \textbf{0.791}  & \textbf{0.794}  \\ \midrule[0.8pt]
\end{tabular}
}
\caption{{ Ablation study on both CANDI and LPBA40 datasets. The best performance is highlighted in bold font. We use the abbreviation  'Seg.' and 'Reg.' for Segmentation and Registration respectively. Measures are averaged over all regions using the $Dice$ coefficient. We denote the baseline as ’B’, patchwise method as ‘PW’, contrastive loss as ‘CL’, and window/shifted-window multi-head self-attention method as “SW”.}} \vspace{-0.5cm}
\label{ablation_study_results}
\end{table}

\section{Discussion}
\label{sec:discussion}
{{From the experiments, we observe several strenghts on our model. Firstly, our hypothesis that  patch embeddings are
a more meaningful representation is supported by our experiments. We observed that our technique has a significant (statistically) performance than the current SOTA techniques. \textit{What is the intuition behind our technique?} The spatial structure of the patch that allows capturing not only global but, more importantly, also local anatomical representations. These local and global representations are reflected in capturing fine-grained  details and, therefore, helping the registration to be more robust to the changes in the to-be registered images.}
Secondly, we highlight that our model is not a trivial combination between VoxelMorph and SwinTransformer. Literature on ViT and VoxelMorph uses the off-the-shelf ViT for better feature extraction, i.e., directly replacing the CNN encoder of the VoxelMorph with ViT. However, our motivation for using SW MHA is to stitch deformation field patches. We underline that we did not change the original CNN encoder part of VoxelMorph with SW MHA. Instead, we only use the SW MHA (one layer from SwinTransformer) to stitch patches after the encoder/decoder part of VoxelMorph. Because we only use one layer of SwinTransformer, the computational cost  with a negligible increase -- the Flops of the model without SW MHA is 416.04 GFlops, whilst for our model are 416.10 GFlops, which only increases 0.6 GFlops.} 

\section{Conclusion}
\label{sec:conclusion}
{We introduce a novel unified unsupervised framework for image registration and segmentation.  We propose to rethink these tasks from a patch-based perspective and introduce two patch-based strategies. Firstly, we introduce a novel patch-based contrastive strategy to obtain richer features and preserve anatomical details. Secondly, we design a new patch stitching strategy to eliminate any inherent artifact from the patch-based partition. Our intuition behind the performance gain of our strategies is that through patches we capture not only global but also local spatial structures (more meaningful embeddings). We demonstrated that our technique reported SOTA performance for both tasks.}

\section*{Acknowledgements}
LL gratefully acknowledges the financial support from a GSK scholarship and a Girton College Graduate Research Fellowship at the University of Cambridge. AIAR gratefully acknowledges the financial support of the CMIH, CCIMI and C2D3 University of Cambridge.   CBS acknowledges support from the Philip Leverhulme Prize, the Royal Society Wolfson Fellowship, the EPSRC grants EP/S026045/1 and EP/T003553/1, EP/N014588/1, EP/T017961/1, the Wellcome Innovator Award RG98755, the Leverhulme Trust project Unveiling the invisible, the European Union Horizon 2020 research and innovation programme under the Marie Skodowska-Curie grant agreement No. 777826 NoMADS, the Cantab Capital Institute for the Mathematics of Information and the Alan Turing Institute. Authors are grateful with Pietro Liò for the fruitful discussions.

\bibliographystyle{IEEEtran.bst}
\bibliography{egbib}

\end{document}